# Grasp-type Recognition Leveraging Object Affordance

Naoki Wake, Kazuhiro Sasabuchi, and Katsushi Ikeuchi

*Abstract*—A key challenge in robot teaching is grasp-type recognition with a single RGB image and a target object name. Here, we propose a simple yet effective pipeline to enhance learning-based recognition by leveraging a prior distribution of grasp types for each object. In the pipeline, a convolutional neural network (CNN) recognizes the grasp type from an RGB image. The recognition result is further corrected using the prior distribution (i.e., affordance), which is associated with the target object name. Experimental results showed that the proposed method outperforms both a CNN-only and an affordance-only method. The results highlight the effectiveness of linguistically-driven object affordance for enhancing grasp-type recognition in robot teaching.

## I. INTRODUCTION

Robot grasping has been a major issue in robot teaching for decades. Recent research has proposed learning-based systems to achieve end-to-end robot grasping [1]–[7]. These systems aim to estimate the contact points or motor commands from the visual input. However, the desired grasping can differ for the same target object depending on the action goals. Therefore, a robot teaching framework should benefit from recognizing how a demonstrator grasps an object (i.e., grasp type) as an intermediate representation of grasping.

In most cases, an object is associated with particular prime actions or grasp types [8]–[10]. Inspired by these studies, we propose to use a prior distribution of grasp types to improve a learning-based classification with a convolutional neural network (CNN) (Fig. 1). We refer to the prior distribution as an affordance, the concept proposed by Gibson [11]. An affordance is obtained by searching a database by an object name. Although several studies have reported the effectiveness of using multimodal cues for grasp-type recognition [12], [13], the effectiveness of linguistically-driven object affordance is still poorly understood within the context of learning-based recognition. In this study, we attempt to highlight the effectiveness of the proposed method by comparing it with a CNN-only and an affordance-only method. The main contribution of this work is to propose a pipeline to enhance grasp-type recognition by leveraging an object affordance.

## II. METHOD

### A. Data Preparation

A dataset by Bullock et al. [14], which contains first-person images obtained from four workers, was used. The images were labeled with an object name and grasp type. The grasp type was based on a grasp taxonomy by Feix et al. [15]. Among the grasp types in the taxonomy, we focused on

Naoki Wake, Kazuhiro Sasabuchi, and Katsushi Ikeuchi are with Applied Robotics Research, Microsoft, Redmond, WA, 98052, USA (e-mail: naoki.wake@microsoft.com).

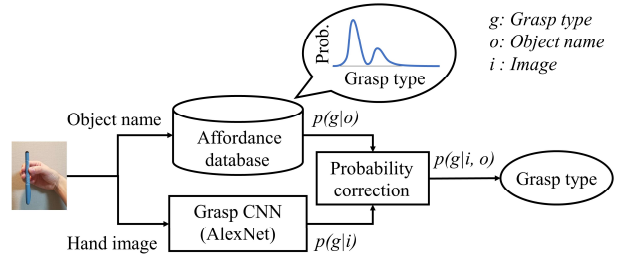

Fig. 1. Proposed pipeline for grasp-type recognition leveraging an object affordance. The pipeline estimates a grasp from a pairing of an object name and an image of the grasping hand.

those labeled frequently in the dataset. Based on the study by Yang et al. [12], we chose the types belonging to power, intermediate, precision, and spherical grasps (Fig. 2a).

We prepared a dataset of grasping images of a machinist (Machinist 1) to learn a CNN-based grasp classifier. A third-party hand detector [16] was applied to crop hand regions from the original images, and detection errors were manually filtered. Fifty images were prepared for each of the four grasp types. Ninety percent of the images were used as a training dataset, and the rest were used as a test dataset. The network was obtained by fine-tuning AlexNet [17]. The prepared dataset was composed of a variety of target objects (Fig. 2b), which may be grasped in several ways.

We prepared an affordance database by calculating a normalized histogram of the labeled grasp types for each object. Fig. 2c shows several examples of object affordances. The affordances were estimated from another machinist (Machinist 2). To avoid contamination of the test and training datasets, we did not include data from Machinist 1.

### B. Fusing CNN and Object Affordance

We formulate a grasp detection by fusing a CNN and an object affordance. Image, object name, and grasp type are denoted as $i$, $o$, and $g$, respectively. We can assume that the output of a CNN and an affordance reflect conditional probability distributions of $p(g|i)$ and $p(g|o)$, respectively (Fig. 1). Further, assuming that $p(i)$ and $p(o)$ are independent, the following equation holds:

$$p(g|i)\, p(g|o) = \frac{p(i|g)\, p(o|g)\, p(g)^2}{p(i)\, p(o)} \qquad (1)$$
$$= p(g|i,o)\, p(g).$$

Hence, a conditional probability distribution $p(g|i,o)$ is estimated by the distributions available: $p(g|i)$, $p(g|o)$, and $p(g)$. Finally, the grasp type is determined as the one that maximizes $p(g|i,o)$.

We evaluated the effectiveness of the proposed pipeline by comparing three methods: the proposed pipeline (i.e., $p(g|i,o)$), a pipeline using only the same CNN (i.e., $p(g|i)$), and a pipeline using only the affordance (i.e., $p(g|o)$). The

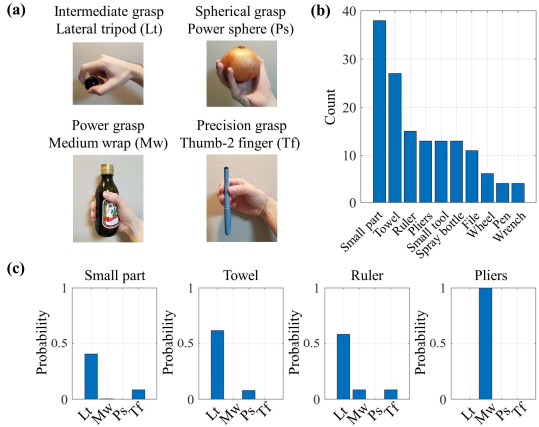

Fig. 2. Grasp types and datasets used for the experiment. (a) The four grasp types focused on in this study. The grasping names are based on [15]. (b) Top ten objects frequently appearing in the training dataset. The dataset was obtained from Machinist 1. (c) Examples of object affordances calculated from Machinist 2. The probability was calculated across 33 grasp types in the original dataset [14].

grasp type that maximizes the probability distribution was chosen.

## III. RESULTS AND DISCUSSION

Table I shows the precision, recall, and overall accuracy of grasp-type recognition using various pipelines. The proposed pipeline demonstrated better performance than the CNN- or affordance-only method in terms of overall accuracy, suggesting the efficacy of combining an object affordance with a CNN. Table II shows examples in which either the CNN- or affordance-only method failed. We observed that the output of the proposed pipeline is reasonable by fusing an output of the CNN and an object affordance. Interestingly, in some cases a novel candidate emerged after the fusing.

For a few images, it was not easy to recognize the grasp type even to human eyes. The difficulties were due to the occlusions of fingers or occlusions of an object in the hand. The use of affordance appears to enhance the CNN-only method in addressing such images with visual ambiguity.

## IV. CONCLUSION AND FUTURE STUDIES

We presented a pipeline that enhances learning-based grasp-type recognition by leveraging an object affordance. The pipeline can be applied to an arbitrary learning-based classification beyond the CNN or dataset considered in this study. Furthermore, the pipeline allows recognizable grasp types to be limited by designing an object affordance $p(g|o)$, based on a concept of task-oriented programming [18].

We believe that the proposed linguistically-driven grasp-type recognition can be usefully employed in a learning from demonstration (LfD) framework, where an object name could be estimated by an accompanying verbal instruction. We are currently testing this hypothesis by integrating the pipeline with an LfD system that we developed in-house [19].

TABLE I    PRECISION (P) AND RECALL (R) FOR EACH GRASP TYPE AND OVERALL ACCURACY

| Grasp type | Lateral tripod | | Medium wrap | | Power sphere | | Thumb-2 finger | | All |
|---|---|---|---|---|---|---|---|---|---|
| | P | R | P | R | P | R | P | R | Acc. |
| CNN | .67 | .40 | .50 | .40 | .50 | .80 | .40 | .40 | .50 |
| Affordance | .26 | 1.0 | 1.0 | .20 | NaN | 0.0 | NaN | 0.0 | .30 |
| CNN+ Affordance | .50 | .80 | 1.0 | .40 | 1.0 | 1.0 | .60 | .60 | .70 |

TABLE II    EXAMPLES OF THE INFERENCE RESULT

| Object name | Towel | Small tool | Rod |
|---|---|---|---|
| CNN | **Power sphere** | Thumb-2 finger | Power sphere |
| Affordance | Lateral tripod | **Lateral tripod** | Lateral tripod |
| CNN+ Affordance | **Power sphere** | **Lateral tripod** | **Thumb-2 finger** |

a. Bold indicates the true grasp types